\documentclass[letterpaper, 10 pt, conference]{ieeeconf}  
\IEEEoverridecommandlockouts                              
\usepackage{graphics} 
\usepackage{subcaption} 
\usepackage{diagbox} 
\usepackage{epsfig} 
\usepackage{mathptmx} 
\usepackage{times} 
\usepackage{amssymb}  

\usepackage{cite}
\makeatletter
\let\NAT@parse\undefined
\makeatother
\usepackage[citecolor=black,colorlinks=true,
            linkcolor=black,
            anchorcolor=black,
            urlcolor=black]{hyperref}

\usepackage{amsmath}
\usepackage{bm}
\usepackage{nomencl}
\makenomenclature

\title{\LARGE \bf
Preserving Relative Localization of FoV-Limited Drone Swarm via Active Mutual Observation
}

\author{Lianjie Guo$^{1,\dag}$, Zaitian Gongye$^{1,2,\dag}$, Ziyi Xu$^{1,2}$, Yingjian Wang$^{1,2}$, Xin Zhou$^{1,2,*}$, Jinni Zhou$^{3}$, and Fei Gao$^{1,2,*}$
\thanks{$^{\dag}$Equal contributors.}
\thanks{$^{1}$Huzhou Institute, Zhejiang University, Huzhou 313000, China.}
\thanks{$^{2}$Institute of Cyber-Systems and Control, College of Control Science and Engineering, Zhejiang University, Hangzhou 310027, China.}
\thanks{$^{3}$The Hong Kong University of Science and Technology (GZ).}
\thanks{$^{*}$Corresponding authors: Xin Zhou and Fei Gao.}
\thanks{E-mail:\{iszhouxin, fgaoaa\}@zju.edu.cn}}

\begin{document}

\maketitle
\thispagestyle{empty}
\pagestyle{empty}

\begin{abstract}

Relative state estimation is crucial for vision-based swarms to estimate and compensate for the unavoidable drift of visual odometry. For autonomous drones equipped with the most compact sensor setting --- a stereo camera that provides a limited field of view (FoV), the demand for mutual observation for relative state estimation conflicts with the demand for environment observation. To balance the two demands for FoV-limited swarms by acquiring mutual observations with a safety guarantee, this paper proposes an active localization correction system, which plans camera orientations via a yaw planner during the flight. The yaw planner manages the contradiction by calculating suitable timing and yaw angle commands based on the evaluation of localization uncertainty estimated by the Kalman Filter. Simulation validates the scalability of our algorithm. In real-world experiments, we reduce positioning drift by up to \textbf{65\%} and managed to maintain a given formation in both indoor and outdoor GPS-denied flight, from which the accuracy, efficiency, and robustness of the proposed system are verified. 
 
\end{abstract}

\section{Introdution}

Micro vision-based aerial swarms have become popular for their low cost, agility, and independence of bulky external sensors. For some swarm missions like formation flight \cite{quan2023robust}, coordinated object handling \cite{zhang2022aerial}, and collaborative mapping and exploration \cite{zhou2023racer}, accurate alignment of the reference frames maintained by each agent is a basic requirement. However, due to the unavoidable drift of vision-based localization, the alignment breaks during mission execution, requiring continuous relative state estimation among vehicles for frame re-alignment. 

To conduct continuous relative state estimation in vision-based swarms, mutual observation is adopted for its environment-independence. In existing methods, mutual observation is achieved through drone detection via onboard cameras. Then relative position can be derived from the drone detection\cite{carrio2018drone,nguyen2020vision,walter2018fast}. Vision-based mutual observation requires drones to be captured by cameras of others, hence some works add extra sensors like multiple fisheye cameras with an omnidirectional sensing range to capture all nearby drones\cite{xu2022omni}. However, for micro aerial vehicles, the limited payload capacity makes it tough to install additional sensors. This paper aims to preserve relative localization with a minimum set of sensors widely applied in autonomous drone navigation: a field-of-view-limited (FoV-limited) stereo camera with an IMU.

\begin{figure}[t]
    \centering
    \includegraphics[scale=0.3]{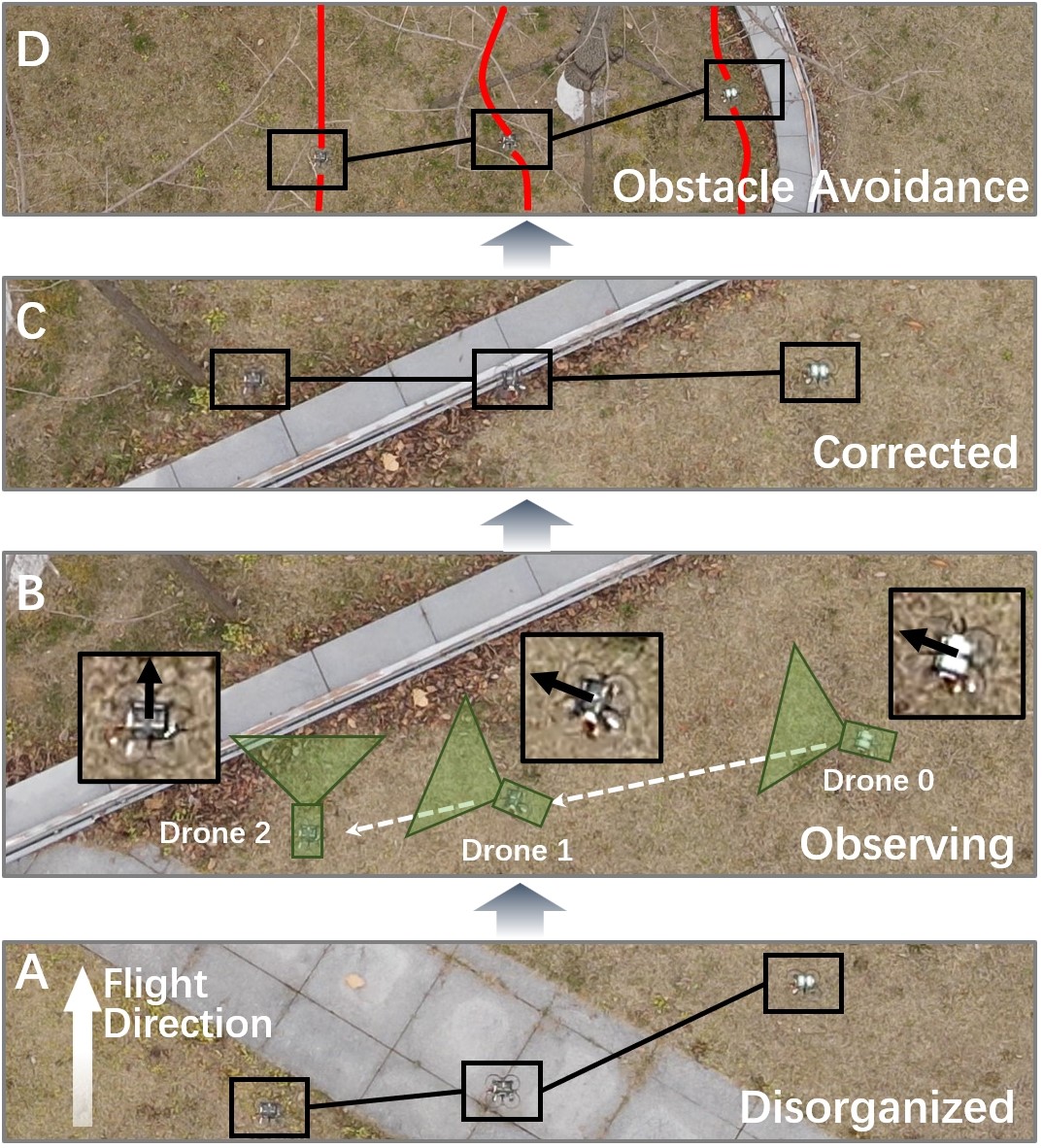}
    \caption{Leverage active mutual observation for localization correction in the field experiment. (A) The drones are disorganized due to the drift of the VIO. (B) The mutual observation tasks are assigned, drone 0 observes drone 1, and drone 1 observes drone 2 (white arrows). The yaw rotation (black arrows) can be seen more clearly in the close-up views. (C) After mutual observation, relative localization is corrected. Drones fly in the predefined line formation. (D) Drones conduct environment observation and deform the formation to avoid the tree obstacles. The red curves are the approximate flight paths.}
    \label{fig: first}
    \vspace{-0.5cm}
\end{figure}

\begin{figure}[t]
    \centering
    \includegraphics[scale=0.35]{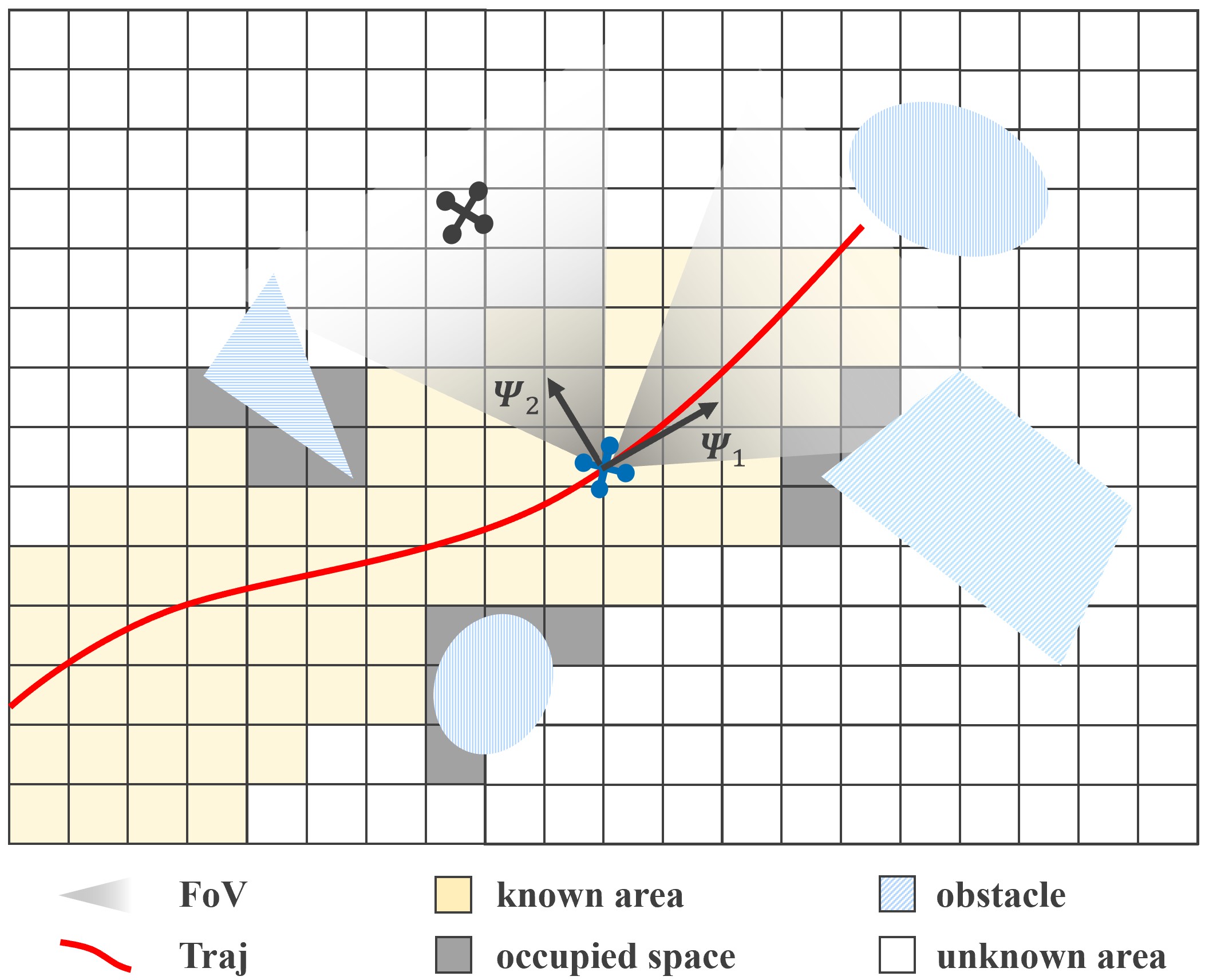}
    \caption{The contradiction between mutual observation and environment observation. If the blue drone chooses environment observation ($\psi_1$), the mutual measurement cannot be obtained due to the black drone out of the FoV of the black drone. If the blue drone chooses mutual observation ($\psi_2$), the drone may collide with the ellipse-shaped obstacle that has not been seen.}
    \label{fig: contradiction}
    \vspace{-0.3cm}
\end{figure}

For FoV-limited swarms, when estimating the relative localization with the only camera, there exists a contradiction between two observations: 1) the environment observation for obstacle avoidance; 2) the mutual observation for relative state estimation (Fig. \ref{fig: contradiction}). To guarantee flight safety, the drone should be oriented toward an area near its future trajectory to get obstacle information. However, mutual observation requires that the drone be oriented toward the others. Commonly, those two requirements cannot be satisfied at the same time due to limited FoV, for example in situations when two drones fly side by side. Faced with this dilemma, some methods \cite{saska2017system} choose to prioritize the demand for environment observation while leaving mutual observation to be satisfied passively and randomly. This means mutual observation is achieved only when a drone happens to fly into the FoV of others, otherwise, the uncertainty grows continuously, and may finally lead to mutual collisions. Without sufficient mutual observation, safety cannot be guaranteed.

To guarantee the flight safety of the swarms, we propose an active localization correction system to balance the two demands in the FoV-limited swarm. The core inside is a yaw planner that decides the time point, duration, target drone, and yaw angle command to perform an observation. Inspired by Roumeliotis et al. \cite{roumeliotis2000distributed}, a Kalman Filter is implemented for swarm state estimation. It takes the position drifts of all the drones as system states, and combines the information of the VIO(Visual-Inertial Odometry) and the relative measurement to calculate corrected localization. The covariance maintained by the Kalman Filter encodes the uncertainty between different drones, which is then used by the yaw planner to determine which kind of observation to choose. A yaw planning process starts from determining a pair of drones with relative localization covariance that satisfies some given criteria. Then the selected drone pair is tasked to perform mutual observation. Before rotating, the required yaw rotation is calculated to make the camera cover a confidence area of the other drone derived from relative localization covariance to ensure the drone is observed. To avoid collision during rotation and ensure that the drone is not blocked by obstacles, we further propose continuous safety and visibility checks. 

In simulation, we tested the scalability and the robustness of the proposed system. In real-world experiments, we conducted indoor and outdoor experiments to verify the capacity to preserve accurate relative localization in both experimental and field environments. The code is released \footnote{\url{https://github.com/ZJU-FAST-Lab/Active-Relative-Localization}} for the reference of the community.

\section{Related Works}

Relative localization approaches for robot swarms can be primarily categorized into two methods: environmental-feature-based and mutual observation. The environmental-feature-based method is commonly employed in Multi-robot SLAM, where agents estimate the relative transformations between robots' map frames by matching common features in their maps. This can be achieved either in a centralized manner\cite{zhang2018cloud,karrer2018cvi} or distributed\cite{lajoie2020door,huang2021disco} fashion. However, due to the exchange of map information and feature-matching requirements, this method is limited in its application to micro-robot swarms due to the increased communication and computational load. On the other hand, mutual observation is a relatively lightweight approach that directly utilizes robot-to-robot range and bearing measurements through methods such as UWB measurement \cite{zhou2022swarm, wu2024scalable, nguyen2023relative}, set markers\cite{cutler2013lightweight,walter2018fast,walter2019uvdar} or reflective tapes detected by LiDAR\cite{zhu2023swarm} to estimate relative localization. 

Utilizing image measurements as a means of mutual observation for vision-based swarms is an intuitive approach. Cutler \cite{cutler2013lightweight} proposes a lightweight solution for estimating range and bearing relative to a known marker, which comprises three IR LEDs in a fixed pattern. Nguyen et al. \cite{nguyen2020vision} propose a visual-inertial multi-drone localization system, and the MAVNet is employed to detect other teammate drones. To further enhance the mutual observation, Xu et al. \cite{xu2020decentralized} introduce a visual-inertial-UWB relative state estimation system that utilizes YOLOv3-tiny for teammate drone detection. In this system, the UWB module is employed as a complementary sensor to provide distance constraints alongside the camera. Nonetheless, all the aforementioned systems may encounter potential failure or drift when the teammate drones are out of the FoV of the camera. This limitation restricts their applicability in scenarios involving formation flight. The objective of this article is to achieve relative localization using the most compact sensor setup commonly employed for aerial navigation: a pair of forward-placed stereo cameras.

\begin{figure*}[t]
    \centering
    \includegraphics[scale=0.43]{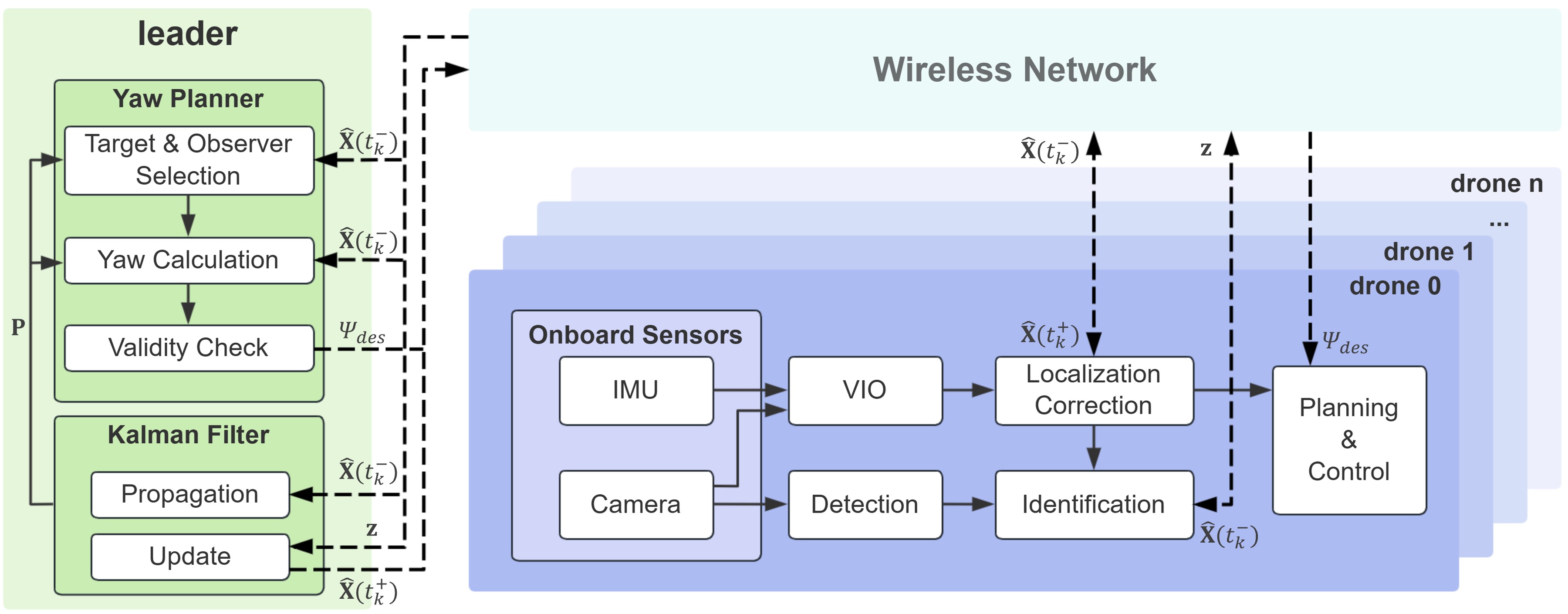}
    \caption{The system architecture of the active localization correction framework. The notation can be referred to the nomenclature in Sec. \ref{notation}. The system consisting of $n$ drones has one leader with a yaw planner and a Kalman Filter deployed. The desired yaw calculated by the leader $\psi_{des}$, the localization result $\mathbf{\hat{X}}$, and mutual observation measurements $\mathbf{z}$ of every drone are transmitted over the wireless network. Based on the covariance $\mathbf{P}$ and the corrected localization $\mathbf{\hat{X}}$ from the Kalman Filter, the yaw planner selects a target and an observer for mutual observation, then calculates the expected yaw $\psi_{des}$, and performs validity checks. After passing the checks, the expected yaw is sent to the drone to execute mutual observation. The drone captured during mutual observation is identified by comparing the estimated position of other drones with the observation measurement. Then the observation measurement is sent to the Kalman Filter for drift estimation. Finally, the estimated drift is used to correct localization information from the VIO.}
    \label{fig: system framework}
\end{figure*}

Upon observing each other, the utilization of acquired relative state measurements varies among different approaches. Early works \cite{roumeliotis2000distributed,martinelli2005multi}, combining bearing and range measurements, use the Kalman Filter(KF) to simultaneously localize a group of mobile robots capable of sensing one another. These methods assume that robots can uniquely identify each of the observed robots in their field of view and measure their relative ranges and bearing vectors, which is frequently not applicable in real-world scenarios. To further address the issue of anonymous observations, Nguyen \cite{nguyen2020vision} extended the coupled probabilistic data association filter to cope with nonlinear measurements. Wang\cite{wang2022certifiably} presented a certifiably optimal algorithm that uses anonymous bearing measurements to formulate a mixed-integer quadratically constrained quadratic problem (MIQCQP) to determine bearing-pose correspondences. However, these approaches still struggle with the challenge of partial observations. In this paper, inspired by \cite{roumeliotis2000distributed}, we propose a novel system where the Kalman Filter is extended to trigger extra observations and incorporate mutual position measurements.

\section{Active Localization Correction System}

The complete active localization correction system is shown in Fig. \ref{fig: system framework}. A yaw planner considering the covariance of the relative localization of every drone is deployed to the leader among the swarm. The yaw planner selects a pair of drones that should conduct mutual observation, then determines the observer and the target in this pair, and calculates the yaw angle required for the observer to trigger the detector, i.e., make the target be in the FoV of the observer. Once the target is captured by the observer, the continuously working detector detects the target and outputs the relative position of the drones in the FoV, and the observed drones are identified using the shared odometry of each drone. The observation measurement is transmitted over the wireless network to the leader of the swarm. Then the localization correction is achieved by a Kalman Filter leveraging the estimated relative position derived from odometries and the measurement from the detector, and the newly estimated drift is sent back to every drone in the swarm.

\section{Localization Correction via Kalman Filter}
Inspired by Roumeliotis et al. \cite{roumeliotis2000distributed}, we apply Kalman Filter which takes the position drifts of all the drones as the states of the system, leveraging its ability to estimate the drift and model the uncertainty of the odometry.

\subsection{Notation}\label{notation}
In order to help understand the localization correction framework, the following notations are defined below and followed by the rest of the paper.

\begin{figure}[t]
    \centering
    \includegraphics[scale=0.4]{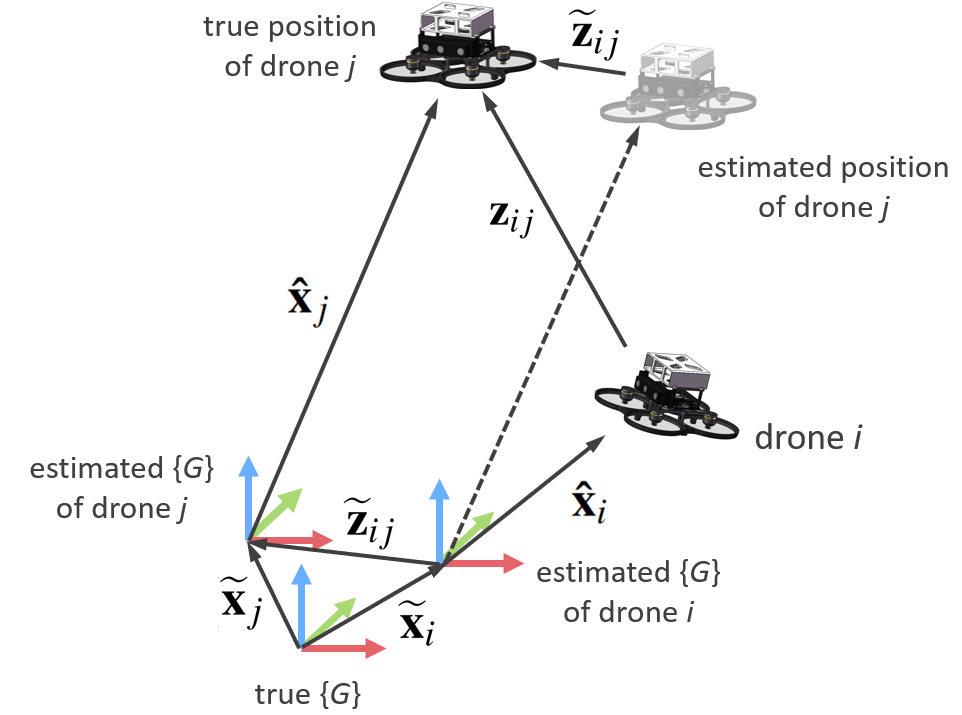}
    \caption{A schematic description of the notation. Frame \{$G$\} means the global frame. Drone $i$ is selected as the observer, and drone $j$ is the target. The estimated global frames of the two drones are not aligned due to the drift $\mathbf{\widetilde{x}}_i$ and $\mathbf{\widetilde{x}}_j$. With the estimated position of itself $\mathbf{\hat{x}}_i$, and the mutual observation measurement $\mathbf{z}_{ij}$, drone $i$ calculates the true position of drone $j$ in the estimated global frame of drone $i$. Then the difference between the true position and the estimated position of drone $j$, $\mathbf{\widetilde{z}}_{ij}$, is the difference between the estimated global frame of two drones.}
    \label{fig: frame}
\end{figure}

\mbox{}
\nomenclature[01]{\( \hat{(\cdot )} \)}{the estimated state}
\nomenclature[02]{\( \widetilde{(\cdot )} \)}{the error state}
\nomenclature[03]{\( (\cdot)(t_k) \)}{the state at discrete-time $t_k$}
\nomenclature[04]{\( (\cdot)(t_k^-) \)}{the state at discrete-time $t_k$ before update}
\nomenclature[05]{\( (\cdot)(t_k^+) \)}{the state at discrete-time $t_k$ after update}
\nomenclature[06]{\( \mathbf{x}_i \)}{the state of drone $i$, $\mathbf{x}_i=[x_i \ y_i \ z_i]^T$ }
\nomenclature[07]{\( \mathbf{X} \)}{the state of all the drones, $\mathbf{X} = [\mathbf{x}_0^T \ ... \ \mathbf{x}_n^T]^T$ }
\nomenclature[08]{\( \mathbf{P}_{ij} \)}{the covariance between drone $i$ and $j$}
\nomenclature[09]{\( \mathbf{P} \)}{the covariance of the whole system}
\nomenclature[10]{\( \mathbf{w}_i \)}{the zero-mean Gaussian noise in the VIO measurement of drone $i$, $\mathbf{w}_i \in \mathbb{R}^{3}$}
\nomenclature[11]{\( \mathbf{W} \)}{the zero-mean Gaussian noise of the VIO measurement from $n$ drones, $\mathbf{W} \in \mathbb{R}^{3n}$}
\nomenclature[12]{\( \mathbf{n} \)}{the zero-mean Gaussian noise in the detector measurement, $\mathbf{n} \in \mathbb{R}^{3}$}
\nomenclature[13]{\( \mathbf{Q} \)}{the corresponding covariance matrix of the noise $\mathbf{W}$}
\nomenclature[14]{\( \mathbf{N} \)}{the corresponding covariance matrix of the noise $\mathbf{n}$}
\nomenclature[15]{\( \{i\} \)}{the frame attached to drone $i$ }
\nomenclature[16]{\( \mathbf{z}_{ij} \)}{the measurement of the relative position of drone $j$ in the frame \{$i$\}  }
\nomenclature[17]{\( \mathbf{R}^i_j \)}{the rotation matrix transforming frame \{$i$\} to frame \{$j$\}}
\nomenclature[18]{\( \mathbf{r}_{ij} \)}{the residual of the relative measurement $\mathbf{z}_{ij}$ }
\nomenclature[19]{\( \mathbf{S}_{ij} \)}{the covariance of the residual $\mathbf{r}_{ij}$ }

\printnomenclature

\subsection{Problem Formulation} 

As shown in Fig. \ref{fig: frame}, considering an aerial swarm system consisting of $n$ drones and each one applying VIO for localization in an initially aligned common frame (the true $\{G\}$), the alignment will be broken due to the drift of VIO (see the estimated $\{G\}$ of drone $i$ and $j$). The localization correction aims to re-align the global frame by estimating the drift, i.e. the error state of all the drones, $\mathbf{\widetilde{X}} = [\mathbf{\widetilde{x}}_0^T \ ... \ \mathbf{\widetilde{x}}_n^T]^T = [\widetilde{x}_0 \ \widetilde{y}_0 \ \widetilde{z}_0 \ \widetilde{x}_1 \ \widetilde{y}_1 \ \widetilde{z}_1 \ ... \ \widetilde{x}_n \ \widetilde{y}_n \ \widetilde{z}_n]^T \in \mathbb{R}^{3n} $.

\subsection{Relative State Estimator}
The Kalman Filter estimates the position drifts of every drone in the swarm $\mathbf{\widetilde{X}}$ in propagation and uses the measurement from mutual observation $\mathbf{z}_{ij}$ for update. 

\subsubsection{Propagation}

We describe the propagation equations for the Kalman Filter using the assumption that the drift of VIO accumulates in its each update, so the discrete-time propagation for drone $i$ is
\begin{equation}\label{single state propagation}
    \mathbf{\widetilde{x}}_i(t_{k+1})=\mathbf{\widetilde{x}}_i(t_{k}) + \mathbf{w}_i(t_k).
\end{equation}

Based on \eqref{single state propagation}, the discrete-time error-state propagation equation for the system is
\begin{equation}
    \mathbf{\widetilde{X}}(t^-_{k+1})=\mathbf{\widetilde{X}}(t^+_{k}) + \mathbf{w}(t_k).
\end{equation}

The covariance propagation equation is
\begin{equation}\label{covariance propagation}
    \mathbf{P}(t^-_{k+1})=\mathbf{P}(t^+_{k}) + \mathbf{Q}(t_k).
\end{equation}

We assume that for any drone $i$ and $j$, $\mathbf{w}_i$ and $\mathbf{w}_j$ are independent, so we have $\mathbf{Q} = diag(\mathbf{Q}_1, \mathbf{Q}_2, ... \mathbf{Q}_n) $, yielding that the covariance between different drones does not change in propagation.

\subsubsection{Update} 

The measurement from the detector is used in the update process. If no observation measurement is received at time $t_k$, We have:
\begin{equation}
\begin{split}
    \mathbf{\widetilde{X}}(t^+_{k})=\mathbf{\widetilde{X}}(t^-_{k}), \quad
    \mathbf{\widetilde{P}}(t^+_{k})=\mathbf{\widetilde{P}}(t^-_{k}).
\end{split}
\end{equation}

When drone $j$ is in the FoV of drone $i$, the detector measures the relative position of drone $j$ in frame \{$i$\}. The measurement is
\begin{equation}
    \mathbf{\widetilde{z}}_{ij}=\mathbf{\hat{x}}_i + \mathbf{z}_{ij} - \mathbf{\hat{x}}_j,
\end{equation}
where $\mathbf{z}_{ij}$ is the detection measurement of drone $i$, $\mathbf{\hat{x}}_i$ is the odometry information of drone $i$, and $\mathbf{\hat{x}}_j$ is the odometry information shared by drone $j$.

And the measurement model of the error state, i.e., the drift, is
\begin{equation}
    \begin{aligned}
        & \mathbf{\widetilde{z}}_{ij}=\mathbf{R}_i^G(\mathbf{\widetilde{x}}_j-\mathbf{\widetilde{x}}_i)+\mathbf{n} \\
        & =\mathbf{R}_i^G[\mathbf{0} \  \mathbf{0} \  ...\mathbf{0}\  \overbrace{-I}^{i-th} \  \mathbf{0} ... \mathbf{0} \  \overbrace{I}^{j-th} \  \mathbf{0} ... \mathbf{0}]\mathbf{\widetilde{X}}+\mathbf{n} \\
        & =\mathbf{H}\mathbf{\widetilde{X}}+\mathbf{n},
    \end{aligned}
\end{equation}
where
$$
\mathbf{H} = \mathbf{R}_i^G[\mathbf{0} \  \mathbf{0} \  ...\mathbf{0}\  -I \  \mathbf{0} ... \mathbf{0} \  I \  \mathbf{0} ... \mathbf{0}],
$$
and $\{G\}$ refers to the global frame.

Then the residual of the relative position measurement and its covariance can be calculated:
$$
\mathbf{r}_{ij} = \mathbf{\widetilde{z}}_{ij} - \mathbf{H}(t_k)\mathbf{\widetilde{X}}(t_k^-),
$$
$$
\mathbf{S}_{ij} = \mathbf{H}(t_{k})\mathbf{P}(t^-_{k})\mathbf{H}^T(t_k)+\mathbf{N}.
$$

Then the Kalman gain, the updated error state, and the updated covariance can be calculated as below:
\begin{equation}
    \mathbf{K}(t_{k})=\mathbf{P}(t^-_{k})\mathbf{H}^T(t_{k}) \mathbf{S}_{ij}^{-1}(t_k),
\end{equation}
\begin{equation}\label{swarm X correction}
    \mathbf{\widetilde{X}}(t^+_{k})=\mathbf{\widetilde{X}}(t^-_k)+\mathbf{K}(t_k)\mathbf{r}_{ij}(t_k),
\end{equation}
\begin{equation}\label{covariance update}
    \mathbf{P}(t_k^+)=\mathbf{P}(t_k^-)-\mathbf{K}(t_k)\mathbf{H}(t_k)\mathbf{P}(t_k^-).
\end{equation}

Observations between drone $i$ and $j$ can help improving drift estimation of other drones like drone $p$ and drone $q$, as long as $p,q$ have observations with $i,j$ before, i.e., one of $\mathbf{P}_{pi,pj,qi,qj} \neq 0$. Then after updating, the covariance between drone $p$ and drone $q$ is
\begin{equation}
\begin{split}
    \mathbf{P}_{pq}(t_k^+) = \mathbf{P}_{pq}(t_k^-) - (-\mathbf{P}_{pi}(t_k^-) + \mathbf{P}_{pj}(t_k^-) ) \\ 
    (\mathbf{R}_i^G \mathbf{S}_{ij}^{-1} {\mathbf{R}_i^G}^T)(\mathbf{P}_{jq}(t_k^-) - \mathbf{P}_{iq}(t_k^-)).
\end{split}
\end{equation}

Then the covariance between each pair of drones are used in active yaw planning to select the observer and the target for mutual observation.

The drift of drone $p$ is corrected as below:
\begin{equation}\label{single state correction}
\begin{split}
    \mathbf{\widetilde{x}}_p(t_k^+) = \mathbf{\widetilde{x}}_p(t_k^-) + (-\mathbf{P}_{pi}(t_k^-) + \mathbf{P}_{pj}(t_k^-)) \\
    {\mathbf{R}_i^G}^T(t_k^-)\mathbf{S}_{ij}^{-1}(t_k^-) \mathbf{r}_{ij}(t_k^-).
\end{split}
\end{equation}

From \eqref{single state correction}, with the existence of the covariance between drone $p$ and the drone involved in the observation (drone $i$ or $j$), the localization drift of drone $p$ is corrected although drone $p$ is not involved in the observation.

The corrected localization of drone $p$ is:
\begin{equation}
    \mathbf{\hat{x}}_p(t_k^+) = \mathbf{\hat{x}}_p(t_k^-) - \mathbf{\widetilde{x}}_p(t_k^-) + \mathbf{\widetilde{x}}_p(t_k^+).
\end{equation}

\section{Active Yaw Planning}

With the covariance maintained by the Kalman Filter, the yaw planner actively reacts to the increasing relative localization uncertainty and constrains it under a certain threshold.

\subsection{Observer and Target Selection}
Without absolute positioning capacities, the drift of the vision-based swarm cannot be completely compensated, while an accurate relative localization can be preserved leveraging mutual observation. So here we use relative localization covariance to determine whether a pair of drones needs mutual observation.

Based on the Kalman Filter, the uncertainty of the relative localization can be calculated by the covariance of the error state in the Kalman Filter. The covariance of the relative localization between drones $i$ and $j$ is
\begin{equation}\label{relative localization covariance}
\begin{split}
    cov(\mathbf{\hat{x}}_i-\mathbf{\hat{x}}_j) & =\mathbf{P}_{ii}+\mathbf{P}_{jj}-\mathbf{P}_{ij}-\mathbf{P}_{ji} \\ & = 
    \begin{bmatrix}
        \sigma^2_{xx} & \sigma^2_{xy} & \sigma^2_{xz} \\
        \sigma^2_{yx} & \sigma^2_{yy} & \sigma^2_{yz} \\
        \sigma^2_{zx} & \sigma^2_{zy} & \sigma^2_{zz} 
    \end{bmatrix}.
\end{split}
\end{equation}

Based on \eqref{covariance propagation} and \eqref{covariance update}, the relative localization covariance increases every time the VIO outputs its localization result, and the covariance is reduced when the mutual observation enhances the correlation between two drones.

Aiming to preserve relative localization, we propose the concept of active mutual observation, which means the drone actively changes its orientation to capture others for mutual observation. The determinant, the largest eigenvalue and the trace can be used to quantify the magnitude of the relative localization covariance, and here we choose the trace $tr_{ij}$ as the criterion for determining whether the drone $i$ and $j$ require mutual observation execution, where
\begin{equation}
    tr_{ij} = tr(cov(\mathbf{\hat{x}}_i-\mathbf{\hat{x}}_j)).
\end{equation}

The proposed yaw planner continuously checks the trace $tr_{ij}$ of the relative localization covariance. For a pair of drones whose $tr_{ij}$ of the relative localization covariance reaches the set threshold, the drones will be assigned the task of executing active mutual observation to reduce the trace. In detail, to determine the roles (observer or target) of two drones, we calculate the expected yaw angle based on the received drone position, and the drone whose difference between the current yaw angle and the expected yaw angle is less than the other's is selected to be the observer, and the other drone is the target.  The process of yaw calculation before determining the roles is detailed below.

\subsection{Yaw Calculation}

With the received positions of a pair of drones, the yaw planner calculates the yaw that allow each drone to see the other. To improve the possibility that the target is captured by the observer, the possible drift will be taken into consideration. 

For the yaw rotation only changes the line of sight in a 2D plane, we consider the drift in the $x$ and $y$ axes. The distribution of the target's location can be described by \eqref{relative localization covariance}, which is elliptical under certain confidence. Specially, the elliptical area can be written as:
\begin{equation}\label{elliptical}
\begin{split}
    & \frac{((x - \hat{x}_j + \hat{x}_i)\cos{\alpha} + (y - \hat{y}_j + \hat{y}_i)\sin{\alpha})^2}{\lambda_1^2} + \\
    & \frac{((y - \hat{y}_j + \hat{y}_i)\cos{\alpha} - (x - \hat{x}_j + \hat{x}_i)\sin{\alpha})^2}{\lambda_2^2} < s,
\end{split}
\end{equation}
where $s$ is determined by the probability of drone presence within the area. We have:
\begin{equation}
\begin{split}
    & \frac{((x - \hat{x}_j + \hat{x}_i)\cos{\alpha} + (y - \hat{y}_j + \hat{y}_i)\sin{\alpha})^2}{\lambda_1^2} + \\
    & \frac{((y - \hat{y}_j + \hat{y}_i)\cos{\alpha} - (x - \hat{x}_j + \hat{x}_i)\sin{\alpha})^2}{\lambda_2^2} \sim \chi^2(2),
\end{split}
\end{equation}
Therefore, the value of $s$ can be determined by referring to a chi-square distribution table. For an area where the target has a 95\% probability of being present, the corresponding value of $ s$ is 5.991.

For the other notation in \eqref{elliptical}, the $\lambda_1$ is the largest eigenvalue of the covariance matrix of the drifts in the $x$ and $y$ axes:
\begin{equation}\label{xy covariance}
    cov_{xy} = 
    \begin{bmatrix}
        \sigma^2_{xx} & \sigma^2_{xy}\\
        \sigma^2_{yx} & \sigma^2_{yy}\\
    \end{bmatrix},
\end{equation}
and $\lambda_2$ is the other eigenvalue, $\alpha$ is the angle between the eigenvector $v_1$ corresponding to $\lambda_1$ and the x-axis.
\begin{equation}
    \alpha = \tan^{-1}(v_1(1) / v_1(0)).
\end{equation}

\begin{figure}
  \centering
  \includegraphics[scale=0.32]{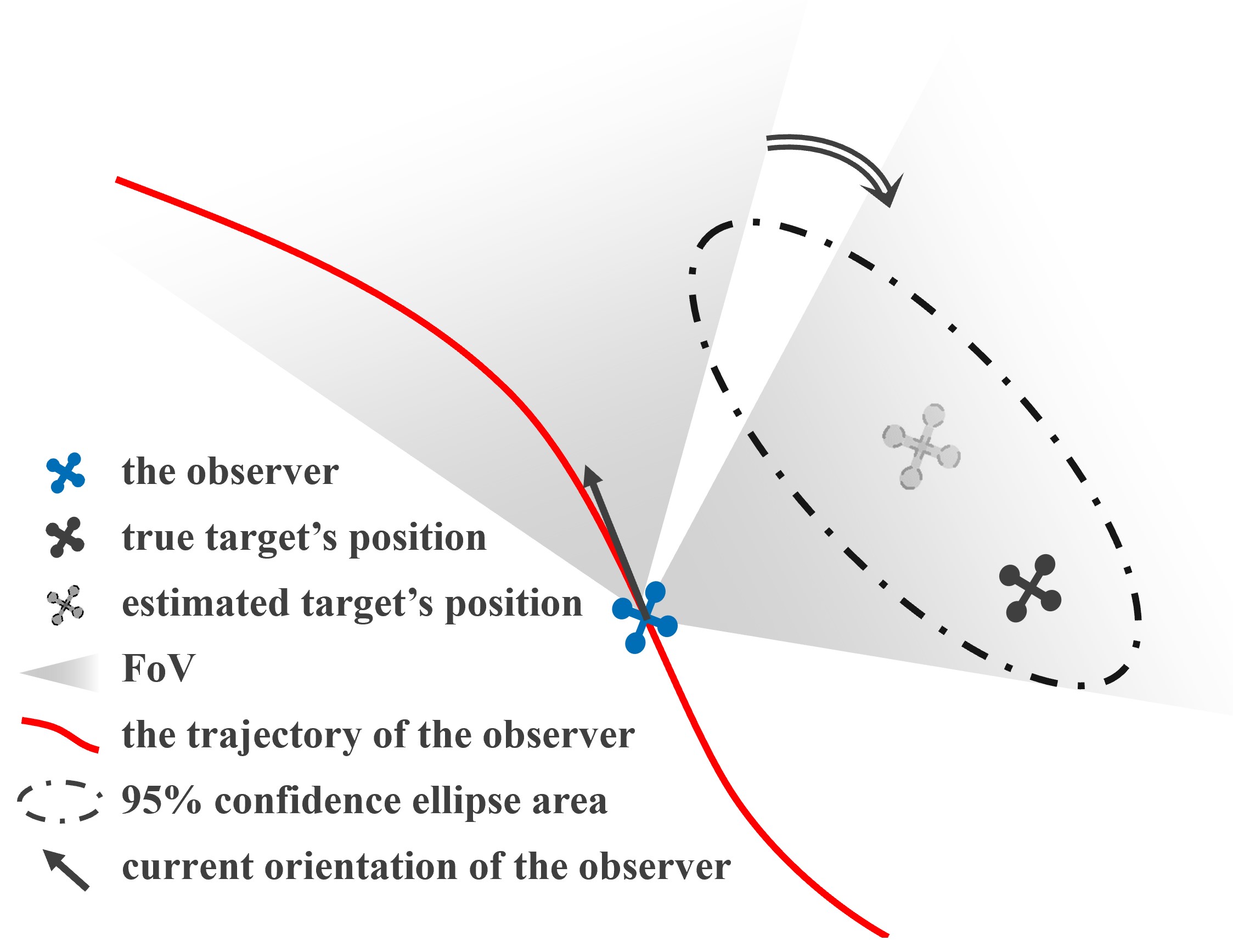}
  \caption{An illustration for yaw calculation. The orientation of the observer is aligned with its velocity direction at first. Because of the drift, The true position of the target is different from the position estimated based on the received position. To improve the possibility of successful observation, the observer should turn right until the right boundary of the FoV tangentially intersects the ellipse}
  \label{cover}
\end{figure}

To guarantee there is a high probability of mutual observation occurring, the line of sight of drone $i$ should sweep across the above-mentioned area as illustrated in Fig. \ref{cover}. Then the expected yaw for the observer to turn to can be calculated. 

\subsection{Validity check}\label{validity check}
Before the active mutual observation starts, safety and visibility checks are performed.

\begin{itemize}
    \item Safety check: 
        Turning causes the drone to lack information about the environment ahead, leading to the possibility that the drone may fly into the unknown environment and crash. With the constant angular velocity $w$ and the expected yaw, the turning time $t_{turn}$ can be easily calculated.
        \begin{equation}
            t_{turn}=\frac{2|\Delta \psi|}{w} = \frac{2|\psi_{des}-\psi_{cur}|}{w},
        \end{equation}
        where $\psi_{des}$ is the desire yaw and $\psi_{cur}$ is current yaw.
    
        The time $t_{out}$ when the drone will fly out of the known map, the current time $t_{cur}$, and the velocity of the drone after mutual observation $v$ can be derived from the executing trajectory. To guarantee flight safety, the drone should be able to stop within the known environment after mutual observation in case there is an obstacle right behind the frontier, so we check if \eqref{safe turning check} holds:
        \begin{equation}\label{safe turning check}
            a_{max}(t_{out}-t_{cur}-t_{turn}) \ge v.
        \end{equation}
    
        If it is not true, the mutual observation task will be canceled.
    \item Visibility check: 
        Limited by the accuracy of the depth information and the camera resolution, the drones that are at distances greater than 3 meters are not considered observation targets. Furthermore, the drones sheltered by the obstacles in the known local map are invisible. The mutual observation task will not be carried out in both conditions.
\end{itemize}

\section{Experiment}
To validate the feasibility and accuracy of the proposed active localization system, we run a series of experiments: 1) scalability evaluation of the centralized Kalman Filter and yaw planner; 2) robustness evaluation to the noise of the VIO and the detection; 3) real-world experiments indoor and outdoor.

\subsection{Detailed System Implementation}

\subsubsection{Detection} As the stereo camera outputs depth and gray-scale images simultaneously, we combine both streams to form dual-channel images, based on which YOLOv8 is adopted for providing real-time object detection on the onboard computer. 

Firstly, we calculate the mean of the 25\% smallest depth pixels within the output bounding box. This approach enable us to disregard the depth of pixels belonging to the background, and instead the depth of pixels constituting the surface of the drone is calculated. Secondly, we compensate the detection measurement with an approximate distance between the surface and the center of the drone, as the distance varies with different observation direction. The compensated value is used as the final depth measurement of the drone.

With the bearing vector provided by the bounding box and the corresponding depth images, the relative position of the drone can be calculated. 

Notably, the detection is executed continuously, which means that for a drone that is not assigned a mutual observation task by the yaw planner, it can detect others that appeared in its FoV, and the mutual measurement is derived and then leveraged by the Kalman Filter. 

\subsubsection{Visual-Inertial Odometry} A GPU-accelerated version of VINS-Fusion\cite{qin2019general} is adopted for ego-state estimation leveraged in the propagation process of the applied Kalman Filter.

\begin{figure}
  \centering
  \includegraphics[scale=0.35]{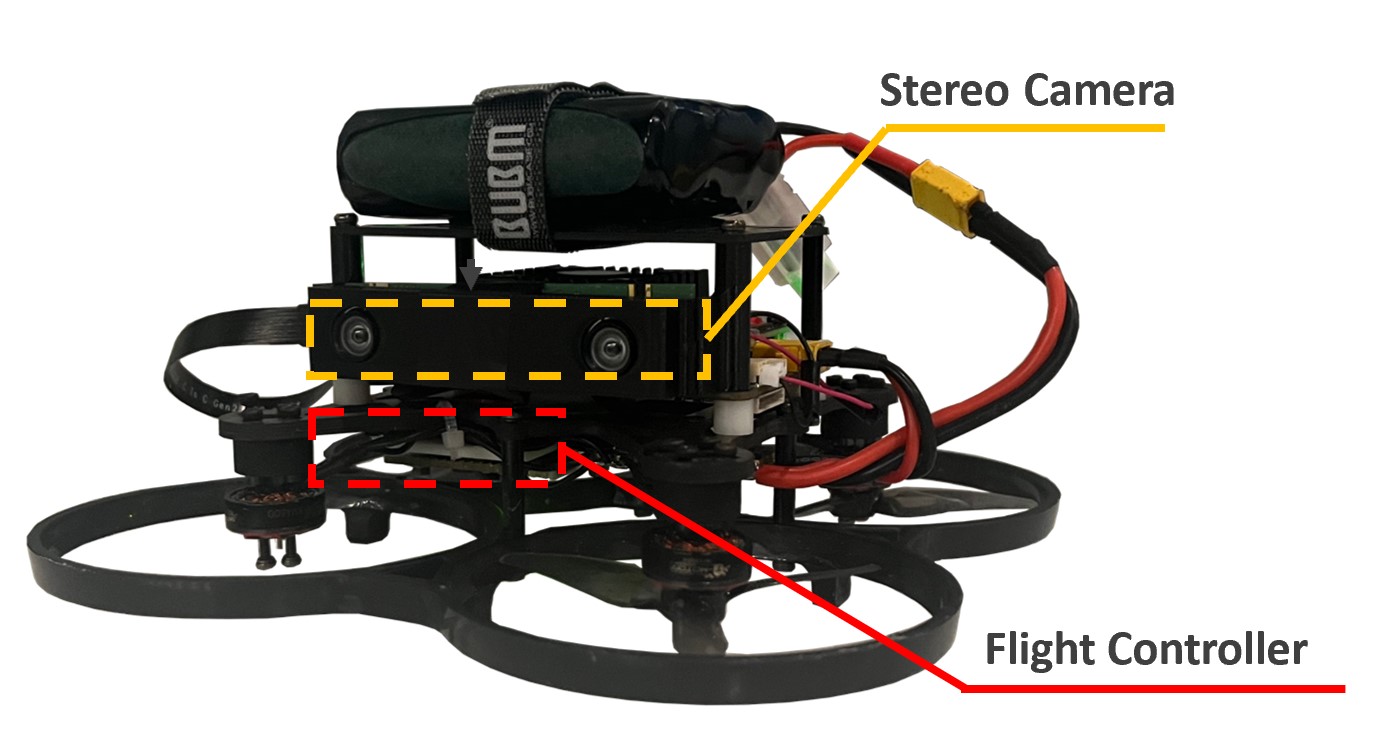}
  \caption{The aerial platform equipped with a minimum set of sensors: a FoV-limited stereo camera with an IMU embedded in the flight controller.}
  \label{platform}
\end{figure}

\subsubsection{Aerial Platform}
Four aerial robots with identical hardware configuration are adopted for the swarm system. Each drone is equipped with a kakute H7 flight controller with an IMU embedded in, an Intel RealSense D430 stereo camera module, and an orin NX onboard computer (Fig. \ref{platform}) .

\subsection{Scalability}

\begin{figure}
  \centering
  \includegraphics[scale=0.28]{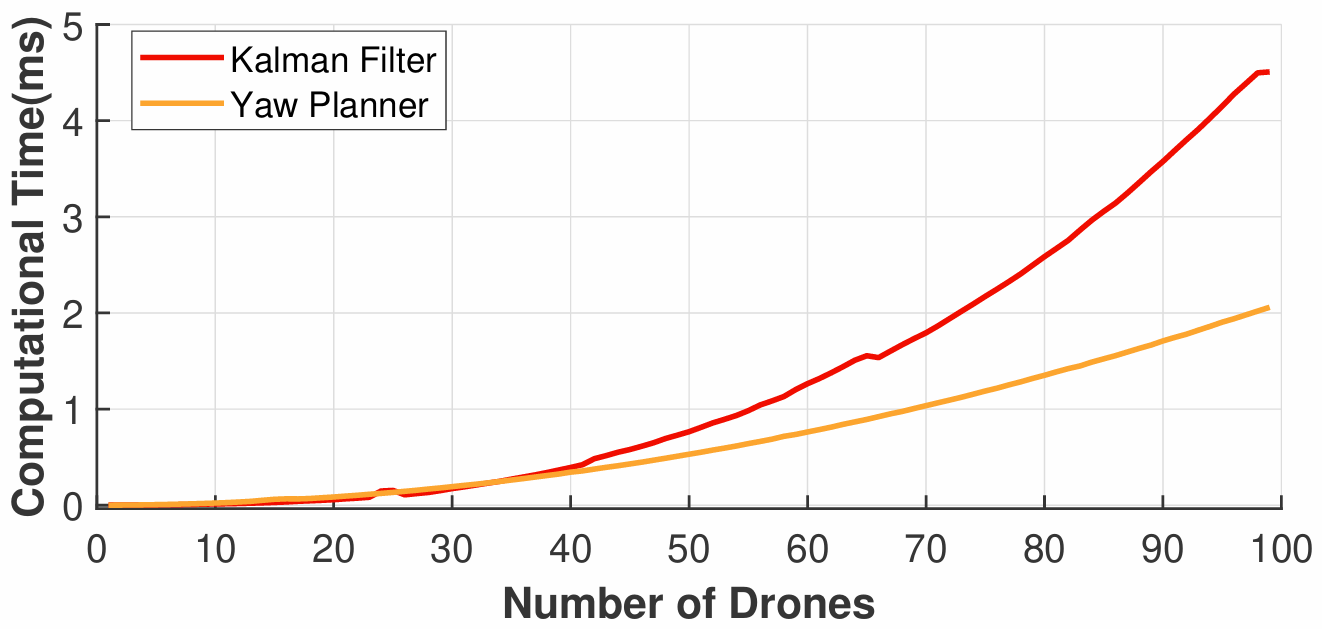}
  \caption{The computation time of the two algorithms.}
  \label{compute_time}
\end{figure}

We test the computation time of the Kalman Filter and the yaw planner in simulation for different quantities of drones in the swarm. Fig. \ref{compute_time} shows how the computation time varies with the number of drones. For the swarm of 100 drones, the Kalman Filter corrects the localization of every drone in 5 ms after receiving mutual measurements, and the yaw planning process can be finished in about 2 ms. The most time-consuming step of yaw planning is the confidence area calculation since it calculates the eigenvalues and eigenvectors of the relative localization covariance of every pair of drones. In real-world scenarios, not every pair needs confidence area calculation for its trace of the covariance does not reach the threshold or the distance between two drones is greater than the valid sensing range. Hence the yaw planning can be executed much faster. In terms of computation time, this evaluation verifies that the method can be extended to a swarm of up to 100 agents without violating the real-time bottom line.

\subsection{Robustness}
To evaluate the robustness of the proposed system, we add different levels of noise into two simulated measurements: 1) the original localization from the VIO; 2) the relative position measurements from the detection. Both of the two kinds of noise are modeled as the zero-mean Gaussian noise. The standard deviations of the three axes position measurements from the VIO is set to be the same value, $\sigma_v$, and the standard deviations of the three axes of the relative position measurements from the detection is also set to be the same value, $\sigma_d$.

Then we change the two deviations to verify the robustness. A four-drone formation flight is executed in the experiment. The total relative pose error (RPE) of the final position of every pair of drones is used to measure the relative localization accuracy. Only positional error is calculated for we only estimate the position drifts:
\begin{equation}
    RPE = \sum_{i=0}^n \sum_{j=0}^n \Vert(\mathbf{x}_i-\mathbf{x}_j) - (\mathbf{\hat{x}}_i-\mathbf{\hat{x}}_j)\Vert.
\end{equation}

\begin{figure}
  \centering
  \includegraphics[scale=0.28]{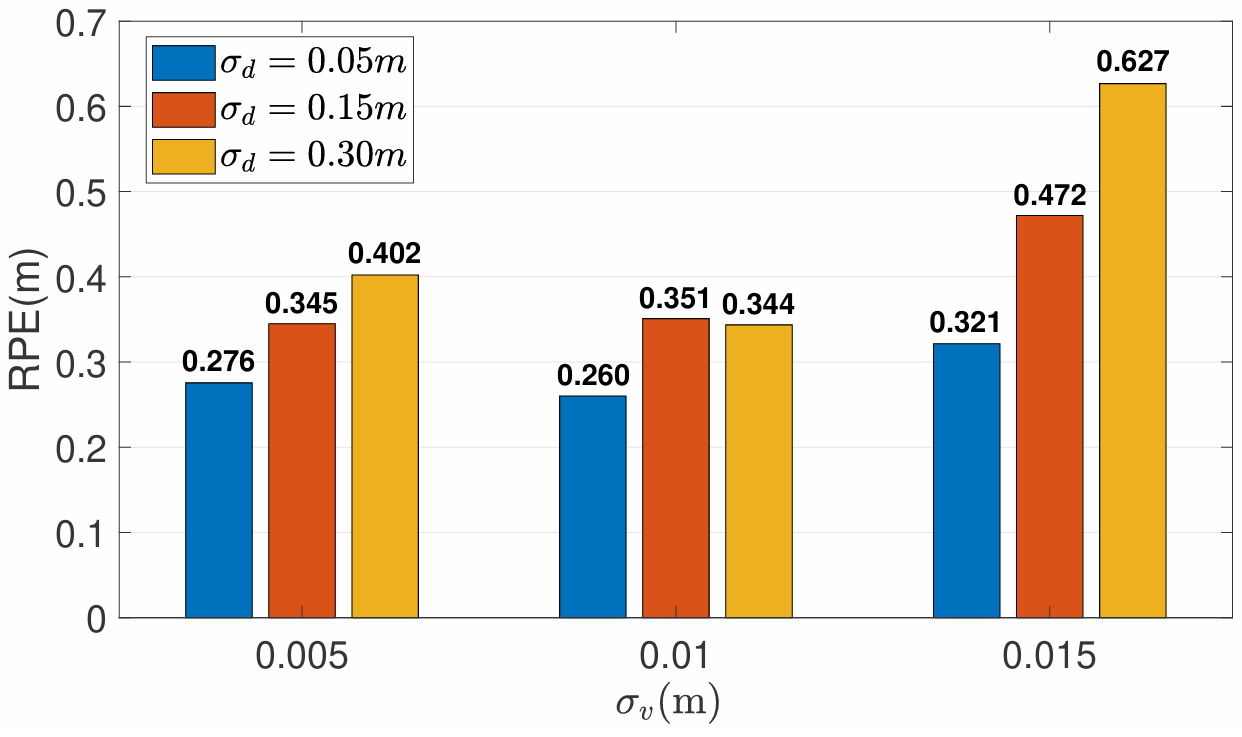}
  \caption{The RPE of the swarm with different levels of noise.}
  \label{robustness}
  \vspace{-0.3cm}
\end{figure}

The RPE under different noise levels is shown in Fig. \ref{robustness}, which reveals the robustness of the proposed system. In the majority of instances, the system's RPE is constrained to a range not exceeding 0.4 meters and shows relatively small variations. The RPE tends to increase only when both measurements are subjected to significant levels of noise. Moreover, the figure shows the capacity of an accurate mutual measurement to constrain the RPE even with substantial noise.

\subsection{Real-world experiments}
\begin{figure}[t]
  \centering
  \includegraphics[scale=0.25]{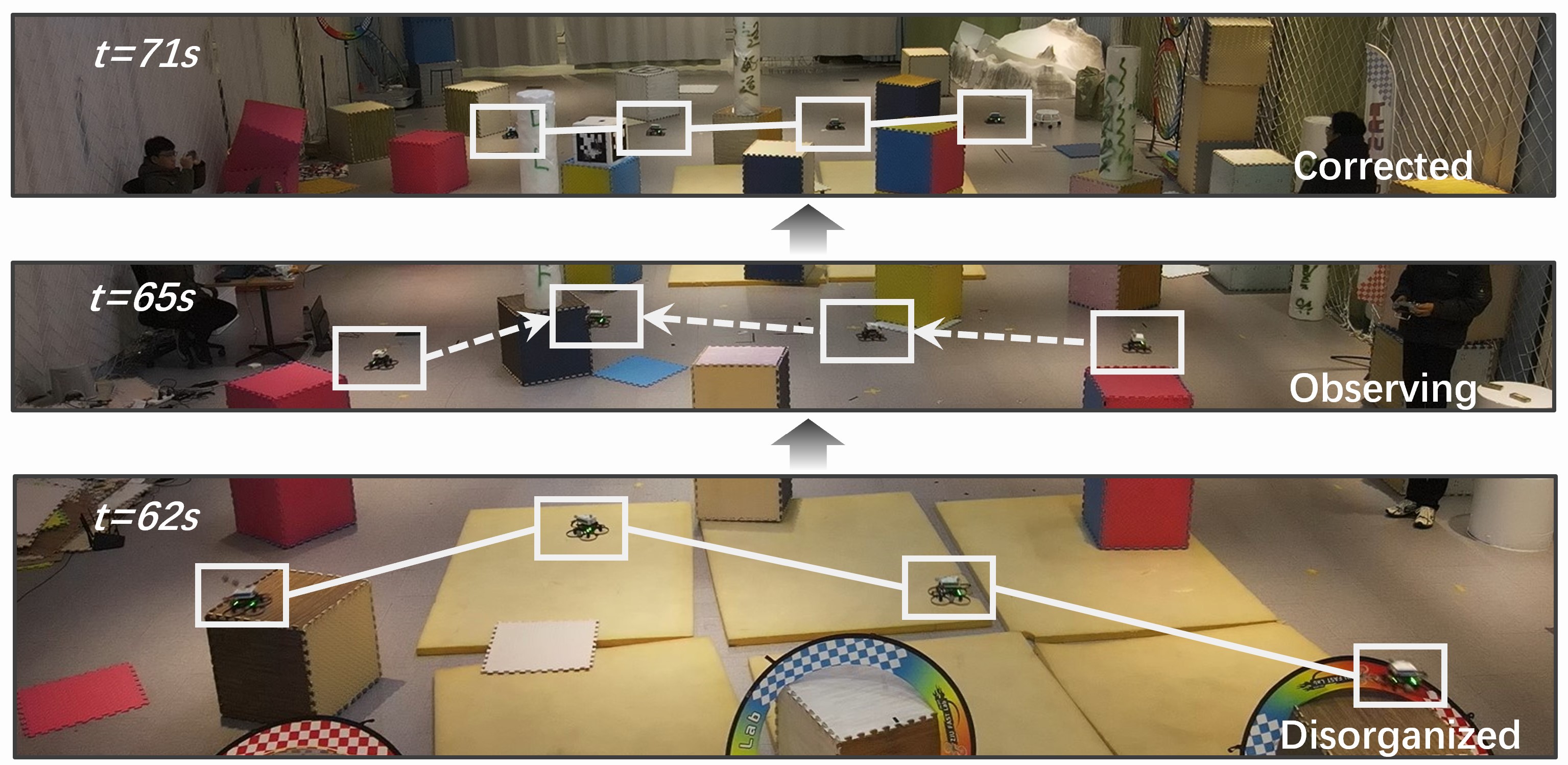}
  \caption{Four drones conduct line formation flight in the indoor experiment.}
  \label{indoor}
\end{figure}

\begin{figure}[t]
  \centering
  \includegraphics[scale=0.3]{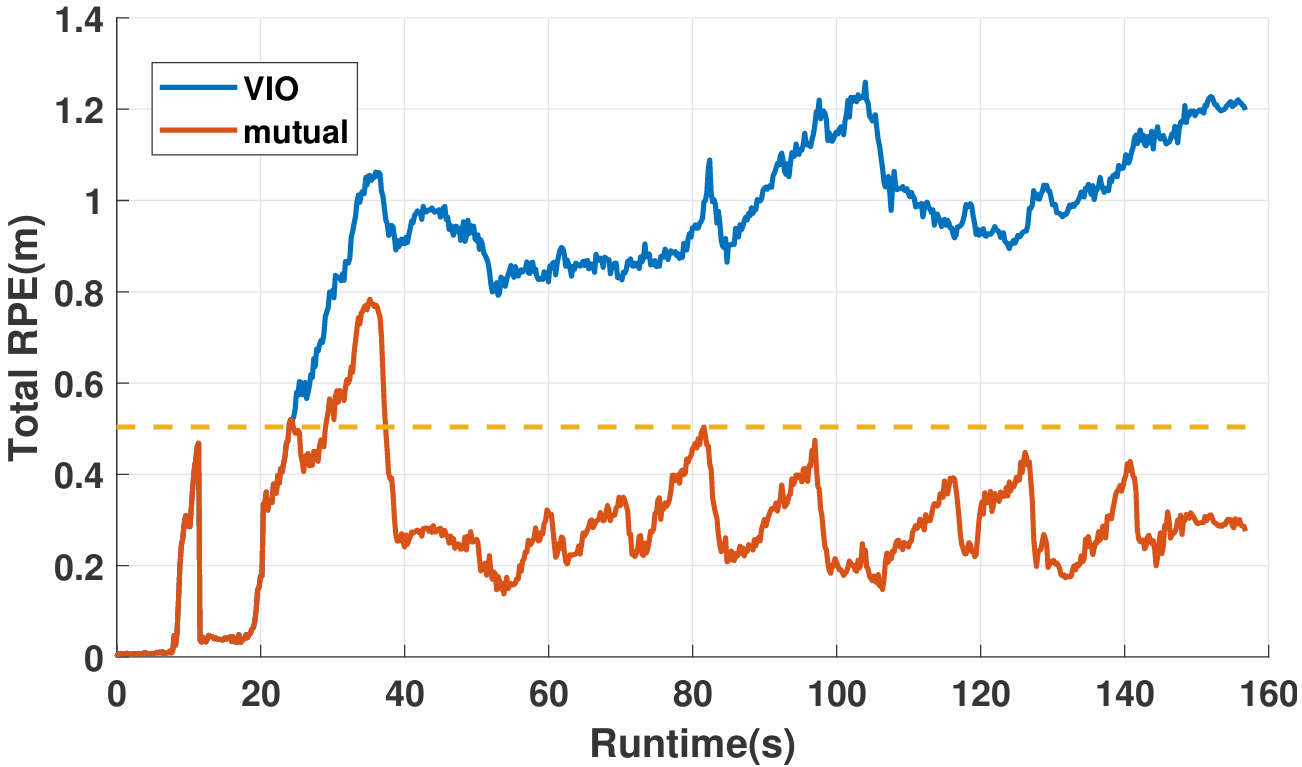}
  \caption{The total RPE of the VIO and the proposed method.}
  \label{total RPE}
  \vspace{-0.3cm}
\end{figure}

For the indoor experiment, the motion capture system is introduced as the ground truth for evaluating the localization accuracy of our method. In the indoor experiment, we deployed four drones, which are expected to fly in a straight line, to execute a formation flight (Fig. \ref{indoor}). Line formation flight is the formation that most requires active observation, for drones cannot capture any other drones without active planning. Fig. \ref{total RPE} shows how the RPE of the estimated position of our method and the VIO changes with time. The RPE is reduced when the four drones are assigned mutual observation tasks at the 40th second, and then the RPE is constrained within a narrow range. 

Besides preserving accurate relative localization, the absolute positioning accuracy has also seen some improvement. Tab. \ref{RMSE} shows the root mean square error (RMSE) of the estimated position of three drones is reduced. However, a relatively more accurate localization of a drone may be suspected to be inaccurate and adjusted because of large drifts of others, which causes the increase in the RMSE of drone $3$.

\begin{table}[t]
\caption{The RMSE of The VIO and The Proposed Method}
\label{RMSE}
\begin{center}
\begin{tabular}{|c|c|c|}
\hline
 & VIO & Proposed Method\\
\hline
Traj Length & \multicolumn{2}{|c|}{91.52m} \\
\hline
RMSE of drone 0 & 0.766m & \textbf{0.347m}\\
\hline
RMSE of drone 1 & 1.205m & \textbf{0.418m}\\
\hline
RMSE of drone 2 & 0.727m & \textbf{0.436m}\\
\hline
RMSE of drone 3 & 0.387m & 0.472m\\
\hline
\end{tabular}
\end{center}
\vspace{-0.3cm}
\end{table}

In the outdoor experiment, as shown in Fig. \ref{fig: first}, three drones start in an area of trees and low bushes in a line. The drones managed to fly through the area safely while maintaining the line formation. The experiment demonstrates that the proposed system can tackle both experimental and field environments. For more details, please view the experimental video.

\section{Conclusion and Future Work}
This paper proposed an active localization correction system for the vision-based FoV-limited swarm. Compared with previous works that utilize visual mutual observation in relative state estimation, we designed an extra module, which calculates appropriate actions for drones to actively execute mutual observation, to solve the contradiction between mutual observation and environment observation that occurs in the FoV-limited swarm. 

In the future, the active mutual observation can be executed via path planning in addition to yaw planning to avoid the loss of environment observation during mutual observation.
\addtolength{\textheight}{0cm}   


\newlength{\bibitemsep}\setlength{\bibitemsep}{0.00\baselineskip}
\newlength{\bibparskip}\setlength{\bibparskip}{0pt}
\let\oldthebibliography\thebibliography
\renewcommand\thebibliography[1]{
	\oldthebibliography{#1}
	\setlength{\parskip}{\bibitemsep}
	\setlength{\itemsep}{\bibparskip}
}
\bibliography{ref}

\end{document}